\documentclass[sn-mathphys-num]{sn-jnl}

\usepackage{graphicx}%
\usepackage{multirow}%
\usepackage{amsmath,amssymb,amsfonts}%
\usepackage{amsthm}%
\usepackage{mathrsfs}%
\usepackage[title]{appendix}%
\usepackage{xcolor}%
\usepackage{textcomp}%
\usepackage{manyfoot}%
\usepackage{booktabs}%
\usepackage{algorithm}%
\usepackage{algorithmicx}%
\usepackage{algpseudocode}%
\usepackage{listings}%
\usepackage{colortbl}
\usepackage[normalem]{ulem}
\useunder{\uline}{\ul}{}
\usepackage{pifont}  

\theoremstyle{thmstyleone}%
\theoremstyle{thmstyletwo}%

\theoremstyle{thmstylethree}%

\begin{document}

\title[Article Title]{WiTUnet: A U-Shaped Architecture Integrating CNN and Transformer for Improved Feature Alignment and  Local Information Fusion}


\author[1]{\fnm{Bin} \sur{Wang}}\email{woldier@foxmail.com}

\author*[1]{\fnm{Fei} \sur{Deng}}\email{dengfei@cdut.edu.cn}

\author[2]{\fnm{Peifan} \sur{Jiang}}\email{jpeifan@qq.com}

\author[2]{\fnm{Suang} \sur{Wang}}\email{wangs@stu.cdut.edu.cn}
\author[1]{\fnm{Xiao} \sur{Han}}\email{hanxiao\_cdut@outlook.com}
\author[3]{\fnm{Zhixuan} \sur{Zhang}}\email{2046236458@qq.com}

\affil*[1]{\orgdiv{College of Computer Science and Cyber Security}, \orgname{Chengdu University of Technology}, \orgaddress{\street{} \city{Chengdu}, \postcode{629100}, \state{ Changchun}, \country{China}}}

\affil[2]{\orgdiv{College of Geophysics}, \orgname{Chengdu University of Technology}, \orgaddress{\street{} \city{Chengdu}, \postcode{629100}, \state{SiChuan}, \country{China}}}

\affil[3]{\orgdiv{College of Mechanical and Vehicle Engineering}, \orgname{Changchun University}, \orgaddress{\street{} \city{Changchun}, \postcode{629100}, \state{Jilin}, \country{China}}}

%


\abstract{Low-dose computed tomography (LDCT) has emerged as the preferred technology for diagnostic medical imaging due to the potential health risks associated with X-ray radiation and conventional computed tomography (CT) techniques. While LDCT utilizes a lower radiation dose compared to standard CT, it results in increased image noise, which can impair the accuracy of diagnoses. To mitigate this issue, advanced deep learning-based LDCT denoising algorithms have been developed. These primarily utilize Convolutional Neural Networks (CNNs) or Transformer Networks and often employ the Unet architecture, which enhances image detail by integrating feature maps from the encoder and decoder via skip connections. However, existing methods focus excessively on the optimization of the encoder and decoder structures while overlooking potential enhancements to the Unet architecture itself. This oversight can be problematic due to significant differences in feature map characteristics between the encoder and decoder, where simple fusion strategies may hinder effective image reconstruction. In this paper, we introduce WiTUnet, a novel LDCT image denoising method that utilizes nested, dense skip pathway in place of traditional skip connections to improve feature integration. Additionally, to address the high computational demands of conventional Transformers on large images, WiTUnet incorporates a windowed Transformer structure that processes images in smaller, non-overlapping segments, significantly reducing computational load. Moreover, our approach includes a Local Image Perception Enhancement (LiPe) module within both the encoder and decoder to replace the standard multi-layer perceptron (MLP) in Transformers, thereby improving the capture and representation of local image features. Through extensive experimental comparisons, WiTUnet has demonstrated superior performance over existing methods in critical metrics such as Peak Signal-to-Noise Ratio (PSNR), Structural Similarity (SSIM), and Root Mean Square Error (RMSE), significantly enhancing noise removal and image quality. The code is available on github \href{https://github.com/woldier/WiTUNet}{https://github.com/woldier/WiTUNet}. }




\maketitle

\section{Introduction}\label{sec1}

In recent years, low-dose computed tomography (LDCT) has gained considerable attention from both the medical community and the public as a promising approach to reduce radiation exposure from X-rays \cite{MarantMicallef2019}. While LDCT technology has mitigated the radiation risks associated with full-dose computed tomography (FDCT) scans to some extent, it has concurrently led to significant degradation in image quality. This reduction in quality arises primarily from the severe noise and artifacts \cite{Tanoue2009} present in LDCT images, posing challenges to accurate disease diagnosis. Consequently, the dual objectives of enhancing image quality while ensuring radiation safety \cite{Shah2008} have become pivotal research areas in medical imaging.

To enhance the quality of LDCT images, noise reduction is a primary and direct strategy. Nonetheless, this remains a formidable challenge due to the ill-posed nature of the LDCT image denoising problem \cite{Huang2021}. To address this, researchers have pursued two main approaches: traditional and deep learning methods (e.g., CNN \cite{Chen2017} \cite{He2016} and Transformer \cite{Wang2023} \cite{Wang2022}). Traditional methods, employing iterative techniques and physical models equipped with specific a priori information, effectively suppress noise and artifacts. However, these methods are often impractical for commercial CT scanners due to hardware limitations and computational demands \cite{Yin2019}. Conversely, With the development of deep learning techniques, CNN-based methods have achieved state-of-the-art performance \cite{Zhang2017}. Notably, Chen et al. \cite{Chen2017} introduced a residual encoder-decoder CNN structure, RedCNN, which achieves effective noise reduction by leveraging residual learning and an architecture similar to Unet \cite{Ronneberger2015}. Additionally, architectures like DnCNN \cite{Zhang2017}, CBDnet \cite{Guo2019}, and NBNet \cite{Cheng2021} have shown robust performance in image processing and are well-suited to handling real noise. DnCNN effectively removes Gaussian noise by utilizing residual learning coupled with batch normalization. Conversely, CBDnet enhances noise level estimation through five convolutional layers, significantly improving the network's generalization performance in noise filtering. NBNet optimizes feature map fusion between the encoder and decoder by integrating a convolutional network within the encoder-decoder skip connections at various downsampling layers, markedly enhancing denoising performance. These deep learning methods have proven to be effective solutions for LDCT image denoising, delivering satisfactory results in practice. They provide crucial technical support for enhancing the quality of medical images and the precision of clinical diagnoses.

Although CNN-based methods have achieved significant advancements in image denoising recently, these methods predominantly focus on using convolutional layers to extract features. However, convolutional layers are limited to capturing local information, leading to CNN-based methods heavy reliance on multiple layers of interconnected convolutional layers to glean non-local information. Additionally, research on encoder-decoder structures remains scant.

In recent years, transformer models \cite{Vaswani2017} have significantly advanced the field of Natural Language Processing (NLP) due to their robust global perception capabilities. These models have also made substantial strides in the domain of computer vision (CV). A notable effort in this integration is the development of the Visual Transformer (ViT) \cite{Dosovitskiy2020} by Dosvitskiy et al., which establishes a linkage between CV and transformers by converting images into patches that serve as tokens for the transformer. Following this, Liu et al. introduced the Swin Transformer \cite{Liu2021}, which augments the contextual sensitivity of each token through patch fusion and a cyclic shift mechanism. However, the large size of LDCT images (typically $512 \times 512$) poses challenges due to the extensive computational demands of global attention. Addressing this, Wang et al. \cite{Wang2023} proposed the CT former, employing an overlapping Window Transformer and Token2Token Dilated strategy within an encoder-decoder architecture for LDCT image denoising, achieving impressive results. The global perception ability of transformers, coupled with the introduction of the attention mechanism, allows these models to overcome the limitations of convolutional layers, which are restricted to local feature perception. This enables the transformer-based methods to enhance global feature extraction and facilitate long-range feature interactions, thereby harnessing more comprehensive information for LDCT image denoising.

Current methodologies, whether based on CNNs or Transformers, exhibit certain limitations. CNN-based models are constrained by their receptive fields, limiting their capacity to extract long-distance contextual information within feature maps. On the other hand, Transformer-based approaches, while potent in global information extraction, tend to overlook local details and suffer from high computational complexity due to their global attention mechanisms. Additionally, methods utilizing either CNNs or Transformers often directly connect feature maps through skip connections in the encoder and decoder, which can lead to suboptimal alignment of context information between these components. This misalignment can adversely affect the reconstruction outcomes, thereby impacting the accuracy of clinical diagnoses.

This paper introduces a novel encoder-decoder architecture that integrates CNNs and Transformers to harness their complementary strengths. The use of nested dense skip pathway in this architecture ensures semantic consistency between the encoder and decoder, enhancing feature map integration. By combining CNNs and Transformers, the network adeptly focuses on global information while maintaining sensitivity to local details, effectively extracting both local and non-local features. Extensive experimental studies and comparisons with existing networks demonstrate that this method significantly reduces noise in LDCT images and enhances image quality.

The main contributions of this paper are as follows:
\begin{enumerate}[(1)]
	\item To address the denoising challenges of LDCT, we introduce a novel encoder-decoder architecture. This architecture features a series of nested dense skip pathway, which are specifically designed to efficiently integrate high-resolution feature maps from the encoder with semantically rich feature maps from the decoder, enhancing information alignment.
	
	\item Recognizing the importance of non-local information for global perception, and acknowledging the high computational demands of traditional global attention mechanisms, we propose a non-overlapping window self-attention module. This module, integrated into our new encoder-decoder architecture, significantly enhances the global perception capabilities for non-local information.
	
	\item To improve the sensitivity to local information within the transformer module, we have developed a new CNN-based block named local image perspective enhancement (LiPe). This block replaces the traditional MLP in the transformers, thereby enhancing local detail capture.
\end{enumerate}  

\section{Related work}\label{}
\subsection{CNN}
Image reconstruction (denoising) aims to restore clarity from a damaged version of an image. In the field of image denoising, a favored solution is to utilize a U-shaped structure with skip connections to capture multi-scale information step by step to construct efficient models \cite{Cheng2021} \cite{Yue2020} \cite{Huang2021}. Zhang et al. \cite{Zhang2021} proposed an image denoising method based on CNN, which enables the network to better learn the noise patterns in the image by introducing a residual learning technique. By introducing the residual learning technique, the network can better learn the noise patterns in the image and thus improve the denoising effect. This method has been experimentally proven to have obvious advantages in denoising performance, showing good robustness to various types and intensities of noise, and bringing a new breakthrough in the development of residual learning image denoising methods. Chen et al. \cite{Chen2017}, a pioneer in the field of image denoising with LDCT, proposed RED-CNN, which incorporates convolution, deconvolution, and skip connections into the encoder-decoder convolutional neural network with U-shaped structure, demonstrating the deep learning method. Yang et al. \cite{Yang2018} used a generative adversarial network (WGAN) with Wasserstein distance to improve the quality of denoised images with the help of perceptual loss. Due to the role of WGAN in generating rich real-world CT images and the role of perceptual loss in improving the quality of denoised images, the model avoids the phenomenon of transition smoothing in the denoised images and preserves the important information in the images. By focusing on how to use the loss function, the network can be better trained to produce images as close to FDCT as possible. Tian et al. \cite{Huang2021} proposed an attention-guided denoising convolutional neural network (ADNet). The network trades off performance and efficiency by removing noise using sparse blocks of dilated convolution and ordinary convolution. The feature enhancement block integrates global and local feature information to solve the problem of hidden noise in complex backgrounds. The authors propose three effective blocks to solve the denoising problem that complement each other: the sparse block improves efficiency but may lead to loss of information, the feature enhancement block compensates for this shortcoming, and the attention block helps to extract noise from complex backgrounds. Chen et al \cite{Cheng2021} proposed NBNet, a deep convolutional neural network utilizing a U-shaped structure. The network is able to learn the noise basis of an image and separate the noise from the image by subspace projection. Its U-shaped structure allows the network to exchange information between the encoder and decoder to better capture subtle features in the image and perform accurate denoising. The experimental results of this study show that NBNet achieves excellent denoising results under various types and intensities of noise, which brings a new and efficient method in the field of image denoising. Huang et al \cite{Huang2021} proposed the DU-GAN method for LDCT image denoising, which utilizes a U-Net based discriminator to learn the differences between denoised and FDCT images in both the image and gradient domains. They also applied another U-Net based discriminator to reduce artifacts and enhance the edges of denoised CT images. Although the use of U-Net as a discriminator and dual domain training strategy in GAN adds some computational cost, the improvement in performance is acceptable.

While the previous discussion highlighted various image denoising methods based on CNN architectures, these methods often encounter specific limitations. First, CNN models are constrained by localized receptive fields, which hinder their ability to capture global information across long distances, particularly in large images. This limitation can adversely affect their performance. Secondly, although the U-shaped structure and skip connections in CNNs aid in the effective transmission of low-level and high-level features, the mechanisms for efficient information exchange between the encoder and decoder are not fully optimized, potentially leading to information loss or redundancy. In contrast, Transformer models excel in global sensing and can capture long-range dependencies without the constraints of local receptive fields, unlike CNNs. Moreover, the attention mechanism within Transformers effectively correlates information across different image locations, facilitating a more integrated approach to combining global and local features.
\subsection{Transformer}
Yang et al. \cite{Yang2022} delved into the CT imaging mechanism and the statistical characteristics of the sinusoidal maps, devising an internal structure loss that encompasses both global and local internal structures to enhance CT image quality. Furthermore, they introduced a sinusoidal map transformation module to more effectively capture sinusoidal map features. By focusing on the internal structure of the sinogram, a feature often overlooked, they significantly reduced image artifacts.

Wang et al. \cite{Wang2022} introduced a novel locally enhanced window (LeWin) transformer block, utilizing a non-overlapping window-based attention mechanism to reduce the computational demands associated with high-resolution feature maps in capturing local information. Employing this module within a U-net architecture, their approach yielded excellent results in image reconstruction. In another contribution, Wang et al. \cite{Wang2023} developed a CT former using an overlapping Window Transformer and Token2Token Dilated strategy within a U-shaped encoder-decoder architecture for LDCT image denoising, achieving remarkable performance. The global perception capabilities of the transformer and the integration of the attention mechanism not only overcome the limitations of convolutional layers' local feature perception but also enhance global feature extraction and facilitate long-range feature interactions, thereby enriching the information used for LDCT image denoising.

While Transformers excel at processing global information, they are comparatively weaker at local perception than CNNs. This limitation can result in suboptimal performance when handling detailed image information. Additionally, much of the existing research has concentrated on enhancing the capabilities of Transformers for image processing tasks, often overlooking the integration with the Unet architecture. This oversight might restrict their effectiveness in simultaneously capturing local and global features. Consequently, it is essential to amalgamate CNNs with Transformers to leverage the strengths of both architectures and to thoroughly investigate the compatibility and effectiveness of the Unet structure within this context.

\section{Method}\label{}
In this section, we first describe the overall architecture of the network. Subsequently, we introduce the Window Transformer (WT) block, which includes the windowed multi-head attention (W-MSA) and the Local Information Perception Enhancement (LiPe). These elements serve as foundational components within both the encoder and decoder. Finally, we discuss the Nested Dense Block, which features nested dense skip pathways that facilitate skip connections between the encoder feature maps and the decoder feature maps.
\subsection{Overall architecture}\label{section3_1}
\begin{figure*}[]
	\centering
	\includegraphics[width=1.0\linewidth]{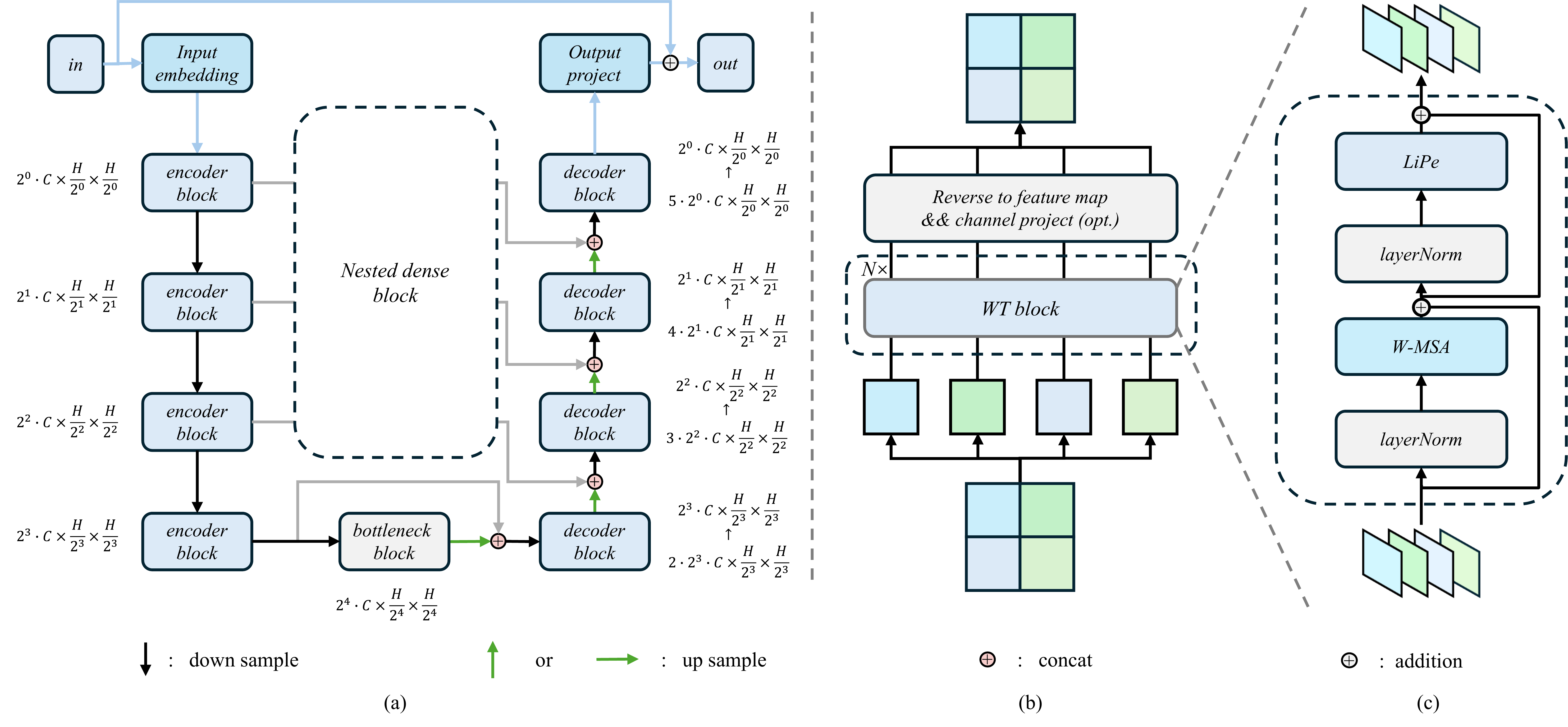}
	\caption{(a) Illustrates the U-shaped network architecture of WiTUnet, featuring encoder and decoder connected by nested dense blocks. (b) Describes the structure of the encoder, bottleneck and decoder, consisting of $N$ WT blocks, and for the decoder, an additional channel projection is depicted.  (c) Details the WT block, highlighting the layers and their functions, including Layer Normalization and Local Image Perception Enhancement (LiPe).}\label{overall_architecture}
\end{figure*}

WiTUnet is presented as a U-shaped network architecture, as illustrated in Figure \ref{overall_architecture} (a). The network consists of encoders, bottleneck, decoders, and an intermediate nested dense block. Specifically, when processing damaged images, namely low-dose computed tomography (LDCT) images represented as $ y\in \mathbb{R}^{1\times H\times W}$, WiTUnet initially employs an input embedding layer, a $3\times 3$ convolutional layer with stride and padding set to one, to transform the raw data into feature maps $x \in \mathbb{R}^{C\times H\times W}$. Following the U-shaped architecture \cite{Ronneberger2015}, these feature maps are processed through $D$ encoder blocks, each comprised of several Window Transformer (WT) blocks, which contain a windowed multi-head self-attention (W-MSA) block. The W-MSA block captures non-local information through its self-attention mechanism and reduces computational complexity using non-overlapping windows, combined with the Local Information Perception Enhancement (LiPe) to precisely capture both local details and global information within the feature maps. Each level of the encoder's output is down-sampled using a $4\times 4$ convolutional layer with a stride of two as input for the next encoder layer, where the feature maps? channel count doubles and their dimensions are halved. After $k$ layers of encoding, the feature mapping is represented as $x_{k,0} \in \mathbb{R}^{2^k\cdot C\times \frac{H}{2^k} \times \frac{W}{2^k} } $, where $k \in [0,D) $.

Between the last encoder layer and the decoder, a bottleneck layer composed of WT blocks is introduced, outputting feature maps as $x_{D,0}\in R^{2^D\cdot C\times \frac{H}{2^D} \times \frac{W}{2^D}}$. This bottleneck is designed to capture global information more effectively. For example, when the encoder depth is sufficient, the deepest feature map dimensions match the window size, allowing the W-MSA to efficiently capture global information, while the LiPe module maintains sensitivity to local details. Thus, the network can perceive global information without the need for cycle shifts \cite{Liu2021}.

In the decoder section, the feature map $x_{D,0}$ undergoes $D$ levels of decoding processing. The decoder structure mirrors the encoder, utilizing stacked WT blocks. Feature maps are upsampled using a $2\times 2$ transposed convolutional kernel with a stride of 2, which halves the channel count and doubles the size. In line with traditional U-shaped network design \cite{Ronneberger2015} \cite{Wang2022}, the decoder receives skip pathways connection from corresponding encoder levels and combines them with the up-sampled inputs from the previous decoder layer. After the introduction of the nested dense block, the channel count of the input feature maps increases, and thus, after combining with the encoder outputs, the input for the $k$-th level of the decoder is \cite{Zhou2018} $x_{k,v;in} \in \mathbb{R}^{(D-K+1)\cdot 2^k\cdot C\times \frac{H}{2^k} \times \frac{W}{2^k} } $. To maintain consistency with traditional U-shaped network decoder outputs, the WT blocks in the decoder adjust the channel count of the output feature maps, resulting in  $x_{k,v;out} \in \mathbb{R}^{2^k\cdot C\times \frac{H}{2^k} \times \frac{W}{2^k} } $, where $k \in [0,D) $ and $v=D-k$. This naming convention aims to simplify the introduction of nested dense blocks in subsequent discussions. The final output after $D$ levels of decoder processing is $x_{0,4;out} \in \mathbb{R}^{2^0\cdot C\times \frac{H}{2^0} \times \frac{W}{2^0} } $, which is then projected onto $r\in \mathbb{R}^{1\times H\times W}$ using a $3\times 3$ convolutional output projection layer with padding and stride set to one, and added to the original LDCT image y to produce the final reconstructed image $\hat{x}=y+r$, where $\hat{x} \in \mathbb{R}^{1\times H \times W}$.

The WiTUnet architecture has been specifically designed to meet the unique requirements of LDCT denoising. Initially, the use of nested dense blocks enhances information flow between different network layers, which aids in reducing noise while preserving crucial image details. Furthermore, the combination of W-MSA modules and LiPe modules allows for the effective capture of global information through the window multi-head self-attention mechanism, and the refinement of local details through local information enhancement components. This integration optimizes the synergy between global and local information, enabling WiTUnet to efficiently balance computational efficiency and reconstruction quality in LDCT image processing. Overall, WiTUnet, with its innovative U-shaped network structure, optimizes information flow and feature integration, offering an effective solution for LDCT denoising. It significantly enhances image quality without sacrificing detail richness, thus providing clearer and more reliable images for clinical diagnosis.

\subsection{Window Transformer block}
Applying Transformers to the domain of LDCT image denoising presents multiple challenges. Firstly, standard Transformers perform global self-attention calculations on all tokens, leading to high computational complexity \cite{Dosovitskiy2020} \cite{Vaswani2017}, particularly problematic given the high-resolution characteristics of LDCT images that increase the dimensions of feature maps and complicate global attention computations. Secondly, while Transformers excel at capturing long-range information through their self-attention mechanism, the preservation of local information is crucial for image denoising tasks, especially in LDCT denoising, which is vital for subsequent clinical diagnosis. Given that Transformers may not capture local details as effectively as CNNs, integrating CNNs to gain more local detail becomes particularly significant.

To address these challenges, this study employs Window Transformer (WT) blocks, as illustrated in Figure \ref{overall_architecture}(c). WT blocks leverage the W-MSA to effectively capture long-distance information while significantly reducing computational costs through a windowing approach. Additionally, LiPe have replaced traditional MLP layers to enhance the WT blocks' capability to capture local information. Figure \ref{overall_architecture}(b) displays the stacking effect of multiple WT blocks. In the decoder, due to the change in the number of channels following the WT blocks, a channel projection is performed post-WT block, whereas this is not done in the encoder and bottleneck layers. Notably, the input from the $(l-1)$-th block, denoted as $X_{l-1}$, undergoes processing through W-MSA and LiPe. The computation within WT blocks can be mathematically expressed as follows:
\begin{equation}
	\label{equation1}
	\begin{aligned}
		X^*_{l-1}=W-MSA(LN(X_{l-1})) + X_{l-1}
	\end{aligned}
\end{equation}
\begin{equation}
	\label{equation2}
	\begin{aligned}
		X_{l}=LiPe(LN(X^*_{l-1})) + 	X^*_{l-1}
	\end{aligned}
\end{equation}
Where $	X^*_{l-1}$and $X_l$ represent the outputs of W-MSA and LiPe, respectively, and LN denotes Layer Normalization \cite{Ba2016}.

\textbf{non-overlapping Window-based Multi Head Self-Attention (W-MSA).} In this paper, we utilize the non-overlapping W-MSA mechanism, which significantly reduces computational complexity compared to the global self-attention mechanism used in vision Transformers. Consider a two-dimensional feature map $X\in \mathbb{R}^{C\times H \times W}$, where $C$, $H$ and $W$ represent the number of channels, height, and width, respectively. We partition $X$ into $N=\frac{H\cdot W }{M^2 }$ non-overlapping windows of size $M\times M$, each of which is flattened and transformed into $X_i\in \mathbb{R}^{M^2\times C}$. Each window is then processed by W-MSA, which, if it has d heads, assigns a dimension $d_k=C/d$ to each head The self-attention computation within each window is described as follows:
\begin{equation}
	\label{equation3}
	\begin{aligned}
		X={X^1,X^2,X^3,\dots,X^N;\ where\ N=\frac{H\cdot W }{M^2 } }
	\end{aligned}
\end{equation}

\begin{equation}
	\label{equation4}
	\begin{aligned}
		U^i_k=Attention(X^iW^Q_k,X^iW^K_k, X^iW^V_k), \ i=1,2, \dots,N
	\end{aligned}
\end{equation}

\begin{equation}
	\label{equation5}
	\begin{aligned}
		\hat{X}_k=\{Y^1_k, Y^2_k,Y^3_k,\dots,Y^N_k\}
	\end{aligned}
\end{equation}

\begin{equation}
	\label{equation6}
	\begin{aligned}
		O=concat(\hat{X}_1,\hat{X}_2,\hat{X}_3,\dots,\hat{X}_d)
	\end{aligned}
\end{equation}
Here,$ W_k^Q$, $W_k^K$, $W_k^V\in \mathbb{R}^{C\times d_k}$ are the projection matrices for the query, key, and value of the k-th head, respectively. $\hat{X}_k$ represents the output from all windows for the $k$-th head. By concatenating the outputs from all heads and applying a linear projection, the final output is obtained. Consistent with previous work \cite{Liu2021} \cite{Shaw2018}, we incorporate relative position encoding into our attention mechanism. Thus, the attention calculation is mathematically defined as:
\begin{equation}
	\label{equation7}
	\begin{aligned}
		Attention(Q,K,V)=SoftMax(\frac{QK^T}{\sqrt{d_k} }+B)V
	\end{aligned}
\end{equation}
Where $B$ represents the relative position encoding bias, obtained from the learnable parameter$ B \in\mathbb{R}^{(2M-1)\times (2M-1)}$ . The W-MSA reduces the computational complexity of global self-attention from $O(H^2 W^2 C)$ to $O(M^2 HWC)$.

\begin{figure}[]
	\centering
	\includegraphics[width=0.5\linewidth]{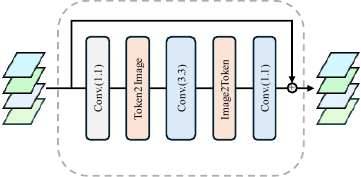}
	\caption{The diagram visualizes the Local Image Perspective Enhancement (LiPe) module, showcasing the sequence of operations including convolutional layers and the transformation from image to token and back, facilitating the feature refinement process within the network architecture.}\label{LiPe}
\end{figure}

\textbf{Local Image Perspective Enhancement(LiPe).} The Feedforward Network (FFN) in standard Transformers, which typically employs a Multi-Layer Perceptron (MLP), has limited capability in capturing local information \cite{Li2021}. For LDCT image denoising, precise restoration of image details is crucial, as these details are vital for accurate disease diagnosis. Due to the properties of CNN convolutional kernels, CNNs exhibit greater sensitivity to local information. To address the shortcomings of the MLP, we have adopted a strategy from prior research  \cite{Li2021} \cite{Sandler2018} \cite{Yuan2021}  and replaced the traditional FFN with a convolutional block. As depicted in the Figure \ref{LiPe}, we initially use a linear projection to increase the channel count for each window's feature map. Subsequently, we reshape the window back to the original feature map form and apply a $3\times 3$ convolutional kernel to capture local details. Afterward, the feature map is windowed and flattened, and then the channel count is restored to its original dimension through another linear projection. Between each layer, we employ the Gaussian Error Linear Unit (GELU) as the activation function.

\subsection{Nested dense block}
\begin{figure*}[]
	\centering
	\includegraphics[width=1.0\linewidth]{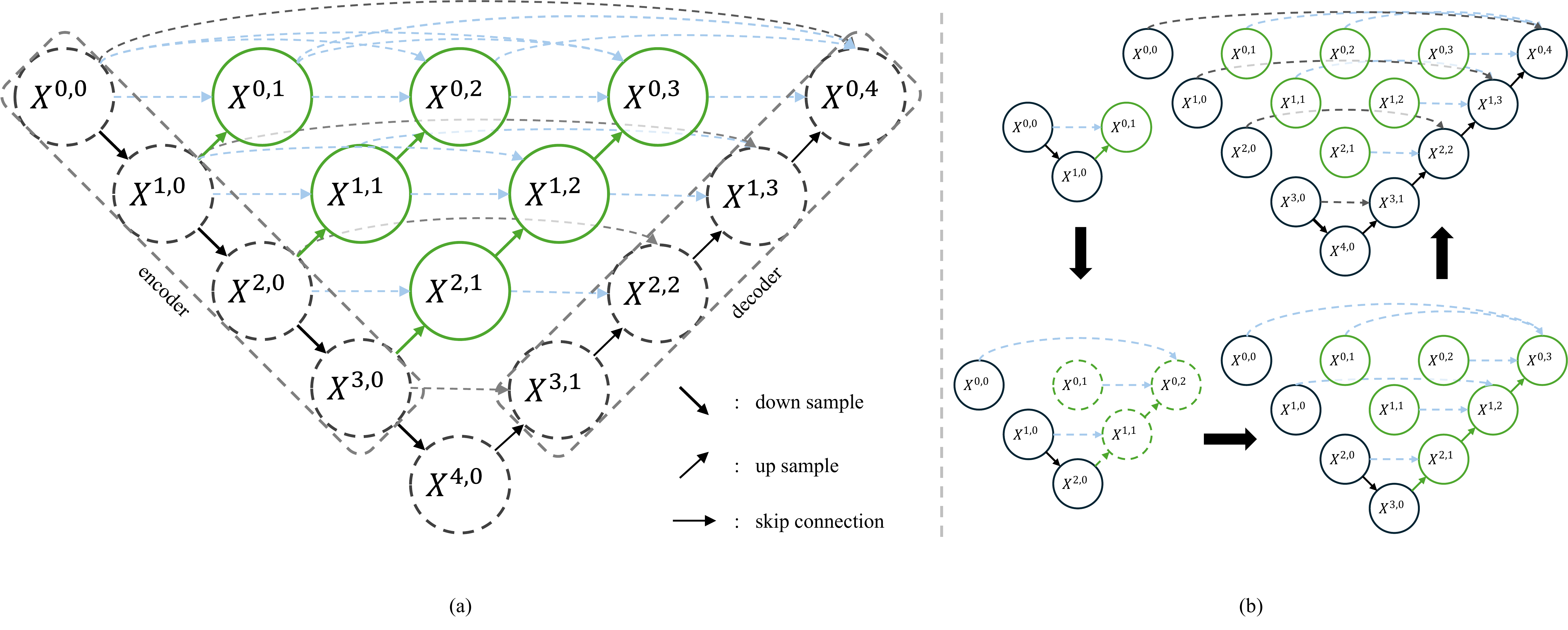}
	\caption{
		(a) Demonstrates the nested dense network structure of WiTUnet, highlighting the complex skip connectiion approach. In the figure, $X_{k,0}$ where $k\in [0,D)$ represents the encoder at different layers, $X_{k,v}$ where $k\in [0,D)$ and $v=D-k$ represents the decoder, and$ X_{D,0}$ is the bottleneck layer. The green and blue arrows show the specially designed jump paths. The redesigned path changes the connection between the codecs, and the encoder's feature map needs to be processed through multiple dense convolutional blocks before it can be passed to the decoder. (b) The computational process at each node is described to help understand the flow and processing of information throughout the network structure.
	}\label{nested_dense_block}
\end{figure*}

Figure \ref{nested_dense_block}(a) presents an alternative perspective of the WiTUnet structure, with a special emphasis on the complex nested dense skip pathway connections within the nested dense block. In the figure \ref{nested_dense_block}, $X_{k,0}$ represents the stages of the encoder, where $k\in [0,D)$ and $X_{k,v}$ denotes the stages of the decoder, where $k\in [0,D)$ and $v=D-k$; $X_{D,0}$ signifies the bottleneck. The remainder of the illustration focuses on the nested dense block mentioned in the article, the primary distinction from the conventional U-Net architecture being the redesigned skip pathways (indicated by green and blue arrows). These redesigned connections alter the interaction between the encoder and the decoder. unlike in the traditional U-Net where feature maps from the encoder are directly passed to the decoder, they are first transmitted through a series of dense convolutional blocks. The number of layers within these blocks correlates with the total number of channels of the feature maps received from the skip connections. Fundamentally, the dense convolutional blocks bring the semantic levels of the encoder feature maps closer to those of the feature maps to be processed in the decoder. We hypothesize that when the encoder feature maps received are semantically similar to the corresponding decoder feature maps, the optimizer faces an easier task in addressing the optimization problem.
Formally, the skip pathways can be represented as follows. Let $x_{k,v}$ denote the output of node  $X_{k,v}$. $k$ represents the depth of downsampling, and $v$ indicates the lateral position on the nested block.

\begin{equation}
	\label{equation8}
	\begin{aligned}
		X_{k,v}=\left\{\begin{matrix}
			X_{k,0}(x_{k-1})&where \  v=0 \\
			X_{k,0}(concat([x_{k,i}]^{v-1}_{i=0},U(x_{k+1,y+1})))& where\ v>0
		\end{matrix}\right.
	\end{aligned}
\end{equation}
It is important to note that $x_{0,0}=X_{0,0}(x_{in} )$. where the notation $U$ denotes the upsampling layer, and the upsampling operation is conducted as described in section 3.1. Typically, nodes at $v=1$ receive inputs from two sources, nodes at $v=2$ from three, and so forth, with nodes at v=i receiving $i+1$ inputs. For a more lucid understanding of the computation involved, Figure \ref{nested_dense_block}(b) visually depicts the computational process at each node.

\section{Experiments and results}\label{}
\subsection{Experiments Detail}
\subsubsection{Datasets}
The dataset publicly released as part of the 2016 NIH-AAPM-Mayo Clinic LDCT Grand Challenge \cite{McCollough2017} has been employed for both training and testing of the model. This dataset comprises pairs of images captured at full dose (acquired at 120 $kV$ and 200 quality reference $mAs$, or $QRM$) and quarter dose (simulated data acquired at 120 $kV$ and 50 $QRM$). It includes abdominal CT scans from ten anonymized patients. For evaluation purposes, data from patient $L506$ were utilized, while the remaining nine patients' datasets were used for model training.

\begin{figure}[]
	\centering
	\includegraphics[width=1.0\linewidth]{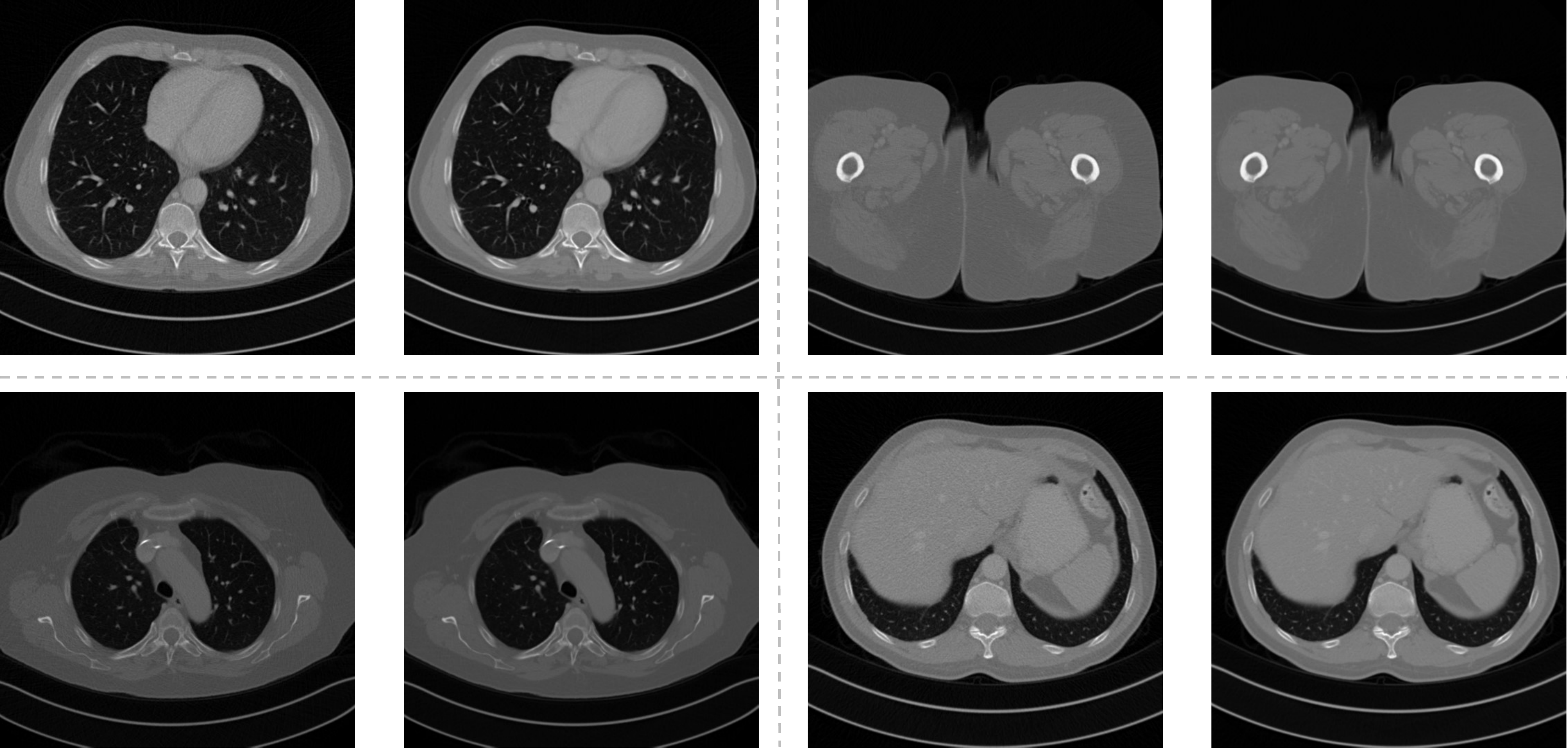}
	\caption{
		Image pair of NIH-AAPM Mayo dataset (partial). 
	}\label{dataset}
\end{figure}

CT scans were originally in DICOM (Digital Imaging and Communications in Medicine) format, with pixel dimensions of $512\times 512$. To expedite processing, we used the Python library pydicom to convert the raw data into NumPy arrays, which were then preprocessed and normalized. Data augmentation techniques were also applied. We generated additional training images by randomly rotating (90, 180, or 270 degrees) and flipping (vertically and horizontally) the original images to further enhance the network's performance. Figure \ref{dataset} showcases a subset of sample pairs from the dataset.

\subsubsection{Train details}
The experimental setup was executed on Ubuntu 18.04.5 LTS with dual AMD EPYC 9654 CPUs. We implemented the WiTUnet model using the PyTorch  \cite{Paszke2019}  deep learning framework, which is widely recognized for its versatility and user-friendly interface. To optimize the training process, the AdamW \cite{Loshchilov2017} optimizer was employed, known for its exceptional performance with large-scale deep learning models. The learning rate was set at $5\times 10^{-4}$, and the betas parameter was in the range of (0.9, 0.99). These parameters were carefully fine-tuned to ensure effective optimization of the network's training process. During training, the network underwent 200 epochs to thoroughly learn the features of the dataset. Furthermore, we set the batch size to one, a common practice to balance memory usage and training efficiency. Notably, to expedite the training phase, computation was carried out on Nvidia RTX 4090 24G GPU $\times$ 4. The parallel processing capabilities of GPUs can significantly accelerate the training speed of deep learning models, enabling researchers to experiment and modify models more swiftly.

\subsection{Results}
\subsubsection{Evaluation measures}
To compare the performance of different denoising approaches, we employed three metrics to ascertain the efficacy of denoising and the quality of LDCT image reconstruction. The evaluation was conducted using the following metrics: peak signal-to-noise ratio (PSNR), structural similarity index (SSIM), and root mean square error (RMSE).

\textbf{Peak Signal-to-Noise Ratio (PSNR).} PSNR is a measurement that defines the ratio of the maximum possible power of a signal to the power of noise that affects its representation. It is frequently used to evaluate the quality of images after denoising. A higher PSNR indicates better quality of the processed image. For a pristine image x and a noise-afflicted image y, PSNR is defined as:
\begin{equation}
	\label{equation9}
	\begin{aligned}
		PSNR=10\times log_{10}(\frac{MAX^2}{MSE})
	\end{aligned}
\end{equation}
Here, MAX represents the maximum pixel value in the image. Note that for CT images where pixel Hounsfield Unit (HU) values may be negative, the actual MAX should be the range between the maximum and minimum pixel values.
\begin{equation}
	\label{equation10}
	\begin{aligned}
		MSE=\frac{1}{mn}\sum_{i=1}^{m} \sum_{j=1}^{n}[x(i,j)-y(i,j)]^2
	\end{aligned}
\end{equation}
\textbf{Structural Similarity Index (SSIM).} SSIM assesses image similarity. It evaluates brightness using the mean, contrast using the standard deviation, and structural similarity using the covariance. For images x and y SSIM is mathematically expressed as:

\begin{equation}
	\label{equation11}
	\begin{aligned}
		SSIM(x,y)=\frac{(2\mu_x \mu_y + C_1 )(2\sigma_{xy}+C_2 )}{(\mu ^2_x + \mu^2_y+C_1)(\sigma^2_x+\sigma ^2_y+C_2)} 
	\end{aligned}
\end{equation}
where  $\mu_x$ $\mu_y$, $\sigma_x^2$ $\sigma_y^2$ represent the means and variances of $x$ and $y$. $\sigma_{xy}$ is the covariance of $x$ and $y$. $C_1$ and $C_2$ are constants used to stabilize the division. The SSIM value ranges from 0 to 1, with higher values indicating greater similarity between $x$ and $y$.

\textbf{Root Mean Square Error (RMSE).} RMSE is sensitive to the error between corresponding pixels in two images. A lower RMSE value indicates higher similarity between the images.

\begin{equation}
	\label{equation11}
	\begin{aligned}
		RMSE= \sqrt{MSE}
	\end{aligned}
\end{equation}

\subsubsection{Comparing method}
For our comparative analysis, we selected several highly regarded and state-of-the-art methods to benchmark against the approach proposed in this paper. The methodologies chosen for this comparison include DnCNN \cite{Zhang2017}, REDCNN \cite{Chen2017}, ADNet  \cite{Tian2020}, NBNet \cite{Cheng2021}, and CTformer \cite{Wang2023}. In the subsequent sections, a comprehensive comparison will be conducted, where the WiTUnet proposed in this study will be rigorously evaluated against these methods.

\subsubsection{Quantitative results}
\begin{figure*}[]
	\centering
	\includegraphics[width=1.0\linewidth]{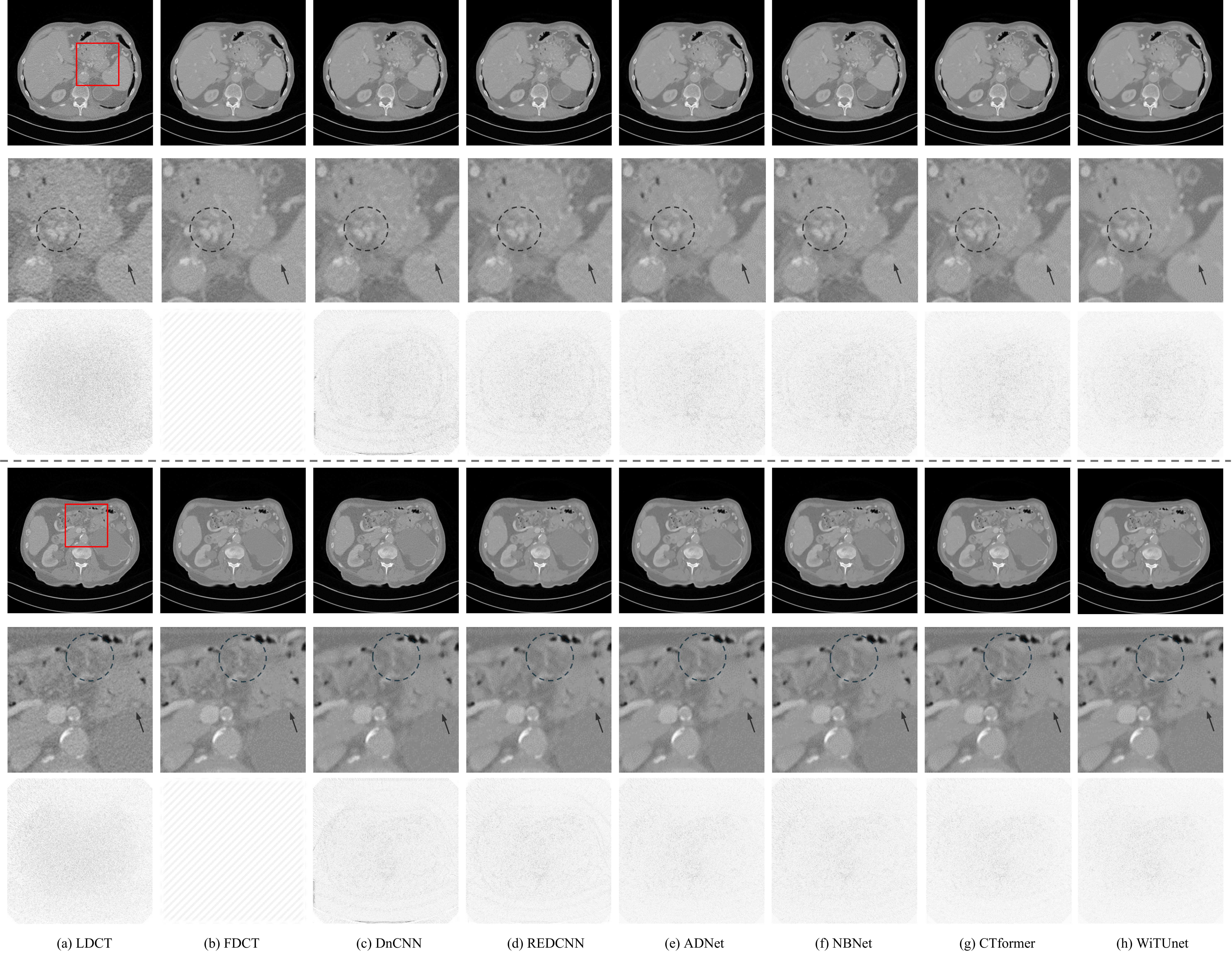}
	\caption{
		(a) LDCT. (b) FDCT. (c) DnCNN. (d) REDCNN. (e) ADNet. (f) NBNet. (g) CTformer. (h) WiTUnet. This figure also includes magnified views of the region of interest (ROI) for each denoising method. To visualize the differences between the denoising results and the FDCT images, differential visualizations are provided, depicting the noise reduction outcome relative to the FDCT. The display window is set to [-160, 240] Hounsfield Units (HU) to enhance the visualization. Additionally, to optimize the visual presentation, the brightness of all visualization results has been adjusted.
	}\label{res}
\end{figure*}
To demonstrate the denoising performance of the method proposed in this study, as well as its ability to suppress noise in LDCT images while retaining rich detail, Figure \ref{res} provides a visualization of denoising results for two slices from the Mayo Clinic 2016 dataset. The regions of interest (ROIs) are demarcated by red dashed rectangles.

The first row, labeled (a) through (h), shows abdominal CT scans from the same anatomical cross-section of a patient, with each column representing a distinct technique or condition. Column (a) shows the original low-dose CT (LDCT) image, which appears noisier with less clarity in the details. Column (b) displays the full-dose CT (FDCT) image, serving as the reference standard with visibly higher clarity and less noise. Columns (c) through (h) present images processed by various denoising algorithms: DnCNN, REDCNN, ADNet, NBNet, CTformer, and WiTUnet respectively. These images exhibit progressive noise reduction and clarity improvement compared to the LDCT image. Beneath the first row, a zoomed-in view of a specific region of interest (ROI) is provided for each technique, allowing for a closer examination of the denoising effects on fine structures within the images. This magnified view facilitates a more detailed comparison of the denoising efficacy, as the subtle textures and contrasts are more discernible. The second row displays the differential images, obtained by subtracting the FDCT images from the denoised results of the respective techniques. These images highlight the differences between the processed images and the FDCT standard, emphasizing areas where noise was reduced or where certain details may have been altered during the denoising process. The variations in grayscale intensity within these differential images correspond to the effectiveness and characteristics of each denoising method. The display window of all images is set to a range of [-160, 240] HU, chosen to best visualize the contrasts and details within the soft tissues and any artifacts present. Additionally, the brightness across all images has been adjusted to maintain a uniform visual standard, ensuring that the comparative differences are not due to variations in image brightness but rather to the denoising algorithms themselves.

Visually, all the compared methods exhibited commendable noise reduction capabilities. However, a detailed examination of the processed images and the magnification of the regions of interest (ROIs) revealed distinct differences between them. For instance, the early denoising network DnCNN, due to its straightforward architecture lacking residual learning, did not achieve denoising performance comparable to more recent models. REDCNN, with its incorporation of an encoder-decoder structure and residual learning, significantly enhanced denoising efficacy. Moreover, ADNet introduced convolutional blocks within its skip connections for better alignment of feature maps between the encoder and decoder, while NBNet, with its attention-guided CNN, bolstered the capture of local details, further improving performance.
Despite the CNN-based models excelling in capturing local information, they exhibit certain limitations in perceiving global information. The CTformer, by adopting a Transformer network and its attention mechanism, improved global information modeling capabilities, demonstrating superior denoising performance relative to other models. However, potentially lacking the benefits of a CNN architecture, CTformer might fall slightly short in capturing fine local details. WiTUnet, merging the global information modeling of window Transformers with the local information capture ability of LiPe and utilizing nested dense blocks for effective feature map alignment, has shown to achieve particularly notable results in denoising.

\begin{table*}[]
	\caption{A comparative analysis of denoising methods delineated by quantitative metrics: SSIM, PSNR, and RMSE, representing the Structural Similarity Index, Peak Signal-to-Noise Ratio, and Root Mean Square Error, respectively. The table ranks DnCNN, REDCNN, ADNet, NBNet, CTformer, and WiTUnet (our proposed method) according to their performance on these metrics. Higher SSIM and PSNR values indicate better image quality and fidelity to the original image, while a lower RMSE signifies closer approximation to the original image with reduced noise. This table provides a succinct numerical summary, reflecting the effectiveness of each method's underlying network architecture in the task of noise reduction while preserving image details.}
	\label{tab:table1}
		\centering
	\begin{tabular}{c|ccc}
		\hline
		METHOD              & SSIM $\uparrow$         & PSNR$\uparrow$          & RMSE$\downarrow$         \\ \hline
		DNCNN \cite{Zhang2017}               & 0.9009        & 31.5689       & 11.1160      \\
		\rowcolor[HTML]{EFEFEF} 
		REDCNN \cite{Chen2017}             & 0.9079        & 32.1245       & 10.0885      \\
		ADNet \cite{Tian2020}               & 0.9115        & 32.7216       & 9.4175       \\
		\rowcolor[HTML]{EFEFEF} 
		NBNet \cite{Cheng2021}               & 0.9092        & 32.4112       & 9.9631       \\
		CTformer  \cite{Wang2023}          & 0.9145        & 31.9098       & 9.2283       \\
		\rowcolor[HTML]{EFEFEF} 
		{\ul WiTUnet(ours)} & {\ul 0.9187} & {\ul 33.2851} & {\ul 8.8434}
	\end{tabular}
\end{table*}

Table \ref{tab:table1} presents the SSIM, PSNR, and RMSE scores for each denoising method. Higher values of SSIM and PSNR indicate superior image quality, while lower RMSE values signify reduced error compared to the image's original state. As an earlier denoising model, DnCNN's performance reflects the limitations of traditional CNN architectures in image denoising tasks. Its relatively simple structure, lacking advanced features such as residual learning or attention mechanisms, resulted in the lowest scores across all three critical metrics. REDCNN, which builds upon DnCNN's framework by incorporating an encoder-decoder structure and residual learning, achieved better performance, especially in terms of image detail preservation (SSIM) and overall image quality (PSNR). This suggests that the encoder-decoder architecture is beneficial for enhanced feature extraction and image reconstruction. ADNet further improved performance by introducing convolutional blocks in its skip connections. This structural optimization helps align feature maps of the encoder and decoder, thus achieving better results on SSIM and PSNR, and notably on minimizing error (RMSE). NBNet, with its attention-guided CNN, strengthened the network's capacity to capture local features. Despite a slight decrease in SSIM, it maintained a relatively low RMSE, demonstrating its potential in detail capturing. CTformer utilized a Transformer-based network architecture, which excels in capturing global dependencies, enhancing the SSIM and PSNR metrics, but did not perform as well on RMSE as WiTUnet, possibly due to its inferior ability to capture fine details compared to CNN-based structures. WiTUnet, integrating the global modeling capabilities of Window Transformers with the local feature capture advantages of CNNs and further augmented by nested dense blocks, showed the best performance across all metrics. Its advanced network structure significantly improved denoising results, particularly in maintaining image structure and reducing reconstruction errors.

\begin{figure*}[!t]
	\centering
		\includegraphics[width=0.317\linewidth]{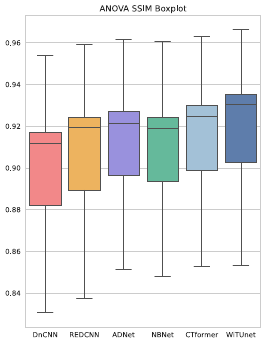}
		\label{anova1}
		\includegraphics[width=0.31\linewidth]{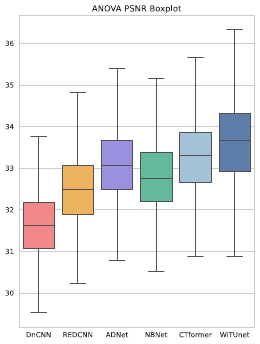}
		\label{anova2}
		\includegraphics[width=0.31\linewidth]{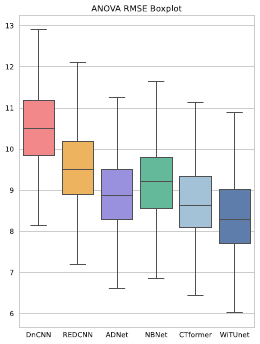}
		\label{anova3}
	\caption{(a) An ANOVA boxplot displaying the distribution of Structural Similarity Index (SSIM) values for six denoising methods, highlighting median, quartiles indicating image quality after denoising. (b) An ANOVA boxplot showing the Peak Signal-to-Noise Ratio (PSNR) performance of the same denoising methods, with higher PSNR values denoting better image restoration quality. (c) An ANOVA boxplot depicting the Root Mean Square Error (RMSE) across the denoising methods, where lower RMSE values represent smaller differences between the denoised and original images, suggesting superior denoising efficacy.}
	\label{anova}
\end{figure*}
Figure \ref{anova} facilitates a comparison of the results from six denoising methods: DnCNN, REDCNN, ADNet, NBNet, CTformer, and WiTUnet, illustrating their statistical performance on three key metrics: SSIM, PSNR, and RMSE. DnCNN, as an early denoising network with a simple structure lacking residual learning, exhibits the weakest performance across all metrics, reflecting its limitations in preserving image quality. In contrast, REDCNN, with its integration of an encoder-decoder structure and residual learning, has significantly enhanced its denoising capabilities, particularly evident in the improved SSIM and PSNR scores. Building on this, ADNet optimizes skip connections with convolutional blocks for better feature map alignment, further advancing denoising effectiveness, especially noted in PSNR enhancement. NBNet, with its attention-guided convolutional neural network, strengthens local information capture, manifested in a slight increase in SSIM, although its improvement in RMSE is not as notable as ADNet?s. CTformer introduces a Transformer network, utilizing its non-local information modeling capacity to demonstrate better denoising outcomes across all metrics, yet it marginally lags behind WiTUnet in RMSE, suggesting room for improvement in local detail restoration. WiTUnet, combining the global modeling strength of window Transformers with local information capture, and augmented by nested dense block alignment, shows the best performance across all evaluation metrics, particularly in structural preservation and error minimization. These results not only showcase the formidable capabilities of advanced denoising network architectures but also reflect the importance of architectural design in influencing denoising quality.

\subsubsection{Ablation study}
\begin{table*}[]
	\caption{Ablation Study Outcomes for WiTUnet. he table showcases the influence of the LiPe module (\ding{172}) and the nested dense block (\ding{173}) on WiTUnet's denoising effectiveness. Metrics reported include SSIM, PSNR, and RMSE. Results indicate that utilizing neither module yields the lowest performance. Implementing the LiPe module alone provides slight enhancements, while employing the nested dense block alone offers further improvements. The incorporation of both modules results in the highest performance across all three metrics, underscoring the combined modules' effectiveness.}
	\label{tab:table2}
	\centering
	\begin{tabular}{l|lllll}
		\hline
		\rowcolor[HTML]{FFFFFF} 
		Name      & \ding{172} & \ding{173} & SSIM $\uparrow$   & PSNR $\uparrow$    & RMSE$\downarrow$   \\ \hline
		WiTUnet   & \ding{56} & \ding{56} & 0.9135 & 28.9212 & 9.3025 \\
		\rowcolor[HTML]{EFEFEF} 
		WiTUnet\ding{172}  & \ding{52} & \ding{56} & 0.9141 & 28.9696 & 8.2559 \\
		WiTUnet\ding{173}  & \ding{56} & \ding{52} & 0.9163 & 29.0179 & 9.2201 \\
		\rowcolor[HTML]{EFEFEF} 
		WiTUnet\ding{172}\ding{173} & \ding{52} & \ding{52} & 0.9187 & 33.2851 & 8.8434
	\end{tabular}
\end{table*}

The ablation study was designed to validate the efficacy of the proposed LiPe module and the nested dense block in enhancing the denoising performance of the network. The impact of each module on denoising was assessed using SSIM, PSNR, and RMSE as performance metrics. Here, symbol \ding{172} represents the LiPe module, while \ding{173} denotes the nested dense block. The absence of \ding{172} implies the utilization of an MLP as the FFN, and the absence of \ding{173} indicates the use of traditional skip connections as found in a conventional Unet architecture.
From our observations, it was found that the network's performance was at its lowest across all metrics when neither \ding{172} nor \ding{173} was employed. Introducing \ding{172} resulted in a noticeable improvement in all evaluative metrics, suggesting an enhanced focus on local information processing by the network. Similarly, the incorporation of \ding{173} led to a higher degree of feature map alignment between the encoder and decoder, thus enabling the network to more effectively reconstruct the FDCT image from the LDCT data. The combined use of both \ding{172} and \ding{173} yielded the highest results across SSIM, PSNR, and RMSE, indicating that the integration of these two modules not only improved the network's local feature extraction capabilities but also facilitated better alignment of information between the encoder and decoder feature maps.

\section{Discussions}
The WiTUnet network has manifested an adeptness in addressing the challenging task of LDCT image denoising, as evidenced by its structural innovations. The nested dense block architecture ingeniously refines information exchange between the encoder and decoder, a feature pivotal in facilitating the optimization process. These advancements underscore the pivotal role of architectural design in medical image processing and set a new benchmark for denoising performance.

The nested dense blocks within the WiTUnet architecture play a critical role in enhancing the alignment of feature maps between the encoder and the decoder. This innovative design element is instrumental in refining the network?s capability to seamlessly integrate and synchronize the information flow across these crucial components of the denoising process. By facilitating more precise alignment, the nested dense blocks ensure that the encoder and decoder work in harmony, which is essential for maintaining the integrity and continuity of the image features. This alignment is vital for recovering high-fidelity details from LDCT images, enabling the network to effectively suppress noise while preserving essential diagnostic information. This sophisticated architectural innovation not only improves the quality of the denoising output but also significantly contributes to the overall accuracy and effectiveness of medical image analysis facilitated by WiTUnet.

Furthermore, WiTUnet's integration of LiPe and W-MSA stands out by enhancing the network's acuity for finer details and a holistic understanding of complex images. This sophistication in processing not only illustrates WiTUnet's superiority in managing intricate content but also its potential for broader clinical application. As the field of LDCT imaging continues to evolve, WiTUnet's contributions offer promising directions for future research and development, particularly in the enhancement of clinical diagnosis through advanced image processing techniques.

\section{Conclusion}
WiTUnet introduces an efficient denoising network structure for LDCT images, distinguished by its innovative nested dense blocks and redesigned skip pathway connections. This architecture not only improves the flow of information between the encoder and decoder but also aligns the semantic levels of the feature maps across different stages, offering the optimizer an advantage in solving optimization challenges. Notably, WiTUnet employs nested skip pathway connections and dense convolutional blocks, considerably enhancing the network's capacity to capture and integrate both global and local features. During the denoising of LDCT images, this structure adeptly balances detail restoration within feature maps with noise suppression, which is of significant importance for increasing the accuracy of subsequent clinical diagnoses. Moreover, with the incorporation of LiPe and W-MSA, WiTUnet further improves its sensitivity to details and global comprehension of complex image content. In summary, WiTUnet demonstrates cutting-edge potential and clinical applicability in the domain of LDCT image processing.

\bibliography{sn-bibliography}

\end{document}